%% file: sdm2024.tex
\documentclass[twoside,twocolumn,leqno]{article}
\usepackage[latin9]{inputenc}
\usepackage[letterpaper]{geometry}
\usepackage{booktabs}
\usepackage{textcomp}
\usepackage{amsmath}
\usepackage{amssymb}
\usepackage{graphicx}
\PassOptionsToPackage{normalem}{ulem}
\usepackage{ulem}
\usepackage[unicode=true,
 bookmarks=false,
 breaklinks=false,pdfborder={0 0 0},pdfborderstyle={},backref=section,colorlinks=false]
 {hyperref}

\makeatletter

\providecommand{\tabularnewline}{\\}

\@ifundefined{date}{}{\date{}}

\usepackage{ltexpprt}
\usepackage{tikz}
\usepackage{booktabs}
\usepackage{multirow}
\DeclareMathOperator{\indep}{\perp\!\!\!\perp}
\DeclareMathOperator{\var}{Var}

\makeatother

\begin{document}
\global\long\def\relatedversion{}%
 
\global\long\def\relatedversion{\thanks{The full version of the paper can be accessed at \url{https://arxiv.org/abs/1902.09310}}}%

\title{{\Large{}Robust Estimation of Causal Heteroscedastic Noise Models}}
\author{Quang-Duy Tran\thanks{Corresponding author, \protect\href{mailto:q.tran@deakin.edu.au}{q.tran@deakin.edu.au}}
\thanks{Applied Artificial Intelligence Institute (A\protect\textsuperscript{2}I\protect\textsuperscript{2}),
\protect \\
Deakin University, Australia} \and Bao Duong\footnotemark[2] \and Phuoc Nguyen\footnotemark[2]
\and Thin Nguyen\footnotemark[2]}

\maketitle

\fancyfoot[R]{\scriptsize{}{Copyright ©\ 2023 by SIAM}\\
{\scriptsize{} Unauthorized reproduction of this article is prohibited}}{\scriptsize\par}




\begin{abstract}
\baselineskip=9pt \input{abs.tex}
\end{abstract}

\section{Introduction\label{sec:Introduction}}

\input{intro.tex}

\section{Related Works\label{sec:RelatedWorks}}

\input{related.tex}

\section{Preliminaries}

\input{preliminaries.tex}

\section{ROCHE: Robust~Estimation of Causal~Heteroscedastic~Noise~Models\label{sec:Methodology}}

\input{method.tex}

\section{Experiments\label{sec:Experiments}}

\input{experiment.tex}

\section{Conclusion\label{sec:Conclusion}}

\input{conclusion.tex}

\bibliographystyle{siam}
\bibliography{references}

\end{document}

%% file: abs.tex
Distinguishing the cause and effect from bivariate observational data
is the foundational problem that finds applications in many scientific
disciplines. One solution to this problem is assuming that cause and
effect are generated from a structural causal model, enabling identification
of the causal direction after estimating the model in each direction.
The heteroscedastic noise model is a type of structural causal model
where the cause can contribute to both the mean and variance of the
noise. Current methods for estimating heteroscedastic noise models
choose the Gaussian likelihood as the optimization objective which
can be suboptimal and unstable when the data has a non-Gaussian distribution.
To address this limitation, we propose a novel approach to estimating
this model with Student\textquoteright s $t$-distribution, which
is known for its robustness in accounting for sampling variability
with smaller sample sizes and extreme values without significantly
altering the overall distribution shape. This adaptability is beneficial
for capturing the parameters of the noise distribution in heteroscedastic
noise models. Our empirical evaluations demonstrate that our estimators
are more robust and achieve better overall performance across synthetic
and real benchmarks.

%% file: intro.tex
The aim of the causal discovery is to uncover the underlying causal
relationships of the data. This task is relevant in various scientific
disciplines such as biology, economics, and sociology~\cite{Pearl_09Causality}.
One foundational challenge in causal discovery involves identifying
the cause and effect between two variables $X$ and $Y$. Randomized
controlled trials (RCT) are considered the ideal solution for determining
causal relationships, particularly in medicine~\cite{Guyon_etal_19Cause}.
However, RCTs require active intervention in variables and observation
of corresponding feedback, making them resource-intensive and sometimes
ethically impractical. To overcome the limitations of RCT, studying
observational data becomes a more challenging yet necessary approach
for identifying causal relationships. 

\begin{figure}[t]
\begin{centering}
\emph{\includegraphics[width=1\columnwidth]{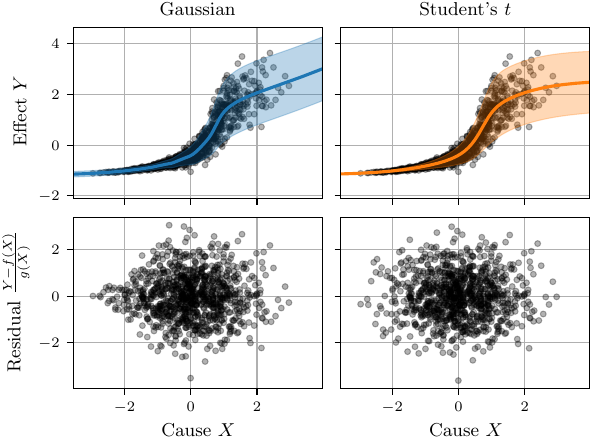}}
\par\end{centering}
\caption{\label{fig:teaser}Heteroscedastic noise models estimated with Gaussian
distribution and Student's $t$-distribution for the likelihood $p\left(Y\mid X\right)$
on pair 82 of the LS-s dataset. The Gaussian estimation fails to capture
the regions with small sample sizes and low variances, which causes
suboptimal fits in these regions. In contrast, estimating with a more
robust Student's $t$-distribution allows the model to better capture
the distribution of the noise. Accordingly, our estimation method
with $t$-distribution can correctly retrieve the residual $N_{Y}=\frac{Y-f\left(X\right)}{g\left(X\right)}$
that is independent of the cause $X$.}
\end{figure}

With solely observational data, determining the causal direction between
$X$ and $Y$ requires prior assumptions of the data-generating process,
which is commonly represented by structural causal models (SCMs)~\cite{Pearl_09Causality}.
In general, an SCM includes a cause variable $X$, an independent
noise term $N_{Y}$ ($N_{Y}\indep X$), and an effect variable $Y$
generated from $X$ and $N_{Y}$ via a function $f$. By restricting
the formulation of $f$, the identifiability or the ability to identify
the causal direction from observational data has been intensively
studied. The earliest and the most comprehensively examined model
is the branch of additive noise models (ANMs), where the noise term
is added after applying the function $f$: $Y:=f\left(X\right)+N_{Y}$.
This model is proven to be identifiable when $f$ is linear and $N$
has a non-Gaussian distribution~\cite{Shimizu_etal_06Linear} or
when $f$ is non-linear~\cite{Hoyer_etal_08Nonlinear}. As the cause
only affects the mean of the noise, ANMs can be estimated easily and
there exist consistent estimators for these models~\cite{Kpotufe_etal_14Consistency}.
Post non-linear noise models (PNLs) are more complex than ANMs where
an additional non-linear invertible function $g$ is applied after
the noise addition: $Y:=g\left(f\left(X\right)+N_{Y}\right)$. Various
works have also been conducted to study the identifiability and estimation
of PNL models~\cite{Zhang_Hyvarinen_09Identifiability,Zhang_etal_15Estimation}.
Heteroscedastic noise models (HNMs) emerge as generalizations of ANMs.
In HNMs, the cause appears in not only the additive form but also
in the multiplicative form: $Y:=f\left(X\right)+g\left(X\right)N_{Y}$.
In addition to identifiability conditions~\cite{Khemakhem_etal_21Causal,Strobl_Lasko_22Identifying,Immer_etal_23Identifiablity},
several methods for predicting the causal direction between two variables
with HNMs~\cite{Tagasovska_etal_20Distinguishing,Khemakhem_etal_21Causal,Strobl_Lasko_22Identifying,Xu_etal_22Inferring,Immer_etal_23Identifiablity}
have been proposed.

Current estimators for HNMs use Gaussian likelihood as the optimization
objective~\cite{Immer_etal_23Identifiablity}. The estimation is
achieved by modeling the mean $\mu\left(X\right)$ and the variance
$\sigma^{2}\left(X\right)$ and maximizing the Gaussian log-likelihood
$\log p\left(Y\mid X\right):=\log\mathcal{N}\left(Y\mid\mu\left(X\right),\sigma^{2}\left(X\right)\right)$.
Although Gaussian distribution is a common choice for the likelihood,
this light-tailed distribution is sensitive to small sample sizes
and extreme values. Figure~\ref{fig:teaser} demonstrates an example
where a region has a small sample size and low variances, the model
fitted with Gaussian likelihood cannot capture the variance correctly.
In this study, we propose ROCHE---a \uline{RO}bust estimation
approach for \uline{C}ausal \uline{HE}teroscedastic noise models
with Student's $t$-distribution using neural networks. The $t$-distribution
is a heavy-tailed distribution that generalizes the Gaussian distribution
with an additional degree of freedom parameter $\nu$. The degree
of freedom $\nu$ controls the heaviness of the tails which improves
the adaptability of the model estimation without changing the general
shape of the distribution significantly. With the robustness in model
estimation, our proposed approach achieves better performance on most
bivariate causal discovery benchmarks with higher stability compared
to models estimated with Gaussian likelihood and other related approaches.

\noindent \textbf{Contributions.} The key contributions of this study
can be summarized as follows:
\begin{enumerate}
\item We propose a robust estimation approach for identifying the causal
direction of heteroscedastic noise models with Student's $t$-distribution.
This approach addresses the limitations of using Gaussian likelihood
and provides a more robust alternative for modeling the noise distribution.
\item We design a framework that incorporates neural networks to robustly
estimate the $t$-distribution likelihood. We introduce essential
constraints that are specifically tailored for the bivariate causal
discovery task, ensuring accurate and reliable estimation results.
\item We demonstrate the effectiveness of our approach by comparing it with
related approaches for both homoscedastic and heteroscedastic noise
models. We evaluate our method on 13 commonly used bivariate causal
discovery benchmarks, showcasing its superior performance and robustness.
\end{enumerate}

%% file: related.tex
Although the bivariate causal discovery or cause-effect inference
from observational data is fundamental and well-defined, this task
still attracts a considerable amount of research attention. Compared
to multivariate settings, the bivariate setting is more limited in
identifying information such as global information about the causal
structures. The recent related literature aiming to infer the cause-effect
relationship between two variables can be categorized into three branches
which will be presented below.

\paragraph{Methods based on cause-effect asymmetry}

The first branch of approaches explores the assumptions of identifiable
structural causal models (SCMs), where the causal models in the anti-causal
directions are proven to be non-existent. Scores for quantifying this
asymmetry in model existence are designed for each type of SCM. Prevalent
scoring choices for determining the causal direction are the log-likelihood
and non-parametric independent test scores such as the Hilbert-Schmidt
Independence Criterion (HSIC)~\cite{Gretton_etal_05Measuring} or
mutual information between the estimated residual and the cause. Various
approaches in this branch has been extended to handle multivariate
data. In homoscedastic/additive noise models where the cause only
is assumed to contribute only to the means, mean regression methods
are being used in many methods for estimating the models~\cite{Shimizu_etal_06Linear,Hoyer_etal_08Nonlinear,Peters_etal_11Identifiability,Peters_etal_14Causal,Peters_Buhlmann_14Identifiability,Buhlmann_etal_14CAM}.
After the estimation step, the maximum likelihood is employed for
inferring the cause and effect in CAM~\cite{Buhlmann_etal_14CAM}.
Correspondingly, RESIT~\cite{Peters_etal_14Causal} also executes
the mean estimation, but an additional step of independence testing
is performed subsequently. 

For heteroscedastic noise models, many works aiming to quantify the
asymmetry have been introduced. CAREFL~\cite{Khemakhem_etal_21Causal}
assumes the invertibility of the functions with the Gaussianity of
the noises and estimates the models with autoregressive normalizing
flows~\cite{Huang_etal_18Neural}. The likelihood of the affine autoregressive
flows is then computed on both training and test data to conclude
the cause-effect direction. HECI~\cite{Xu_etal_22Inferring} divides
the assumed cause $X$ in the several bins and, in each bin, the model
is assumed to have a homoscedastic noise form. The BIC score is computed
in each bin, and the correct causal direction is the one that will
minimize this score. GRCI~\cite{Strobl_Lasko_22Identifying} regresses
the mean and the mean absolute deviation (MAD) with a leave-one-out
cross-validation approach. The residual is computed from the estimated
mean and MAD, and the independence between the residual and the corresponding
cause is assessed by the mutual information. A consistent estimator
and a neural network-based estimator for maximizing the Gaussian log-likelihood
are presented in LOCI~\cite{Immer_etal_23Identifiablity}. Besides
utilizing the log-likelihood as a criterion, from the mean and the
variance parameters of the Gaussian distribution, the residual can
be recovered for the next step of testing the independence with HSIC.
As our proposed approach belongs to this group, ROCHE follows the
same procedure for predicting the causal direction as previous methods.

\paragraph{Methods based on independence between cause and mechanism}

The second branch of methods relies on the postulate of the independence
between the cause and the mechanism. If the causal direction between
$X$ and $Y$ is $X\rightarrow Y$, the marginal distribution of the
cause $P\left(X\right)$ is independent of the conditional distribution
of the effect given the cause $P\left(Y\mid X\right)$ and not vice
versa. IGCI~\cite{Janzing_etal_12Information} considers the case
with low noise levels and formularizes this independence in terms
of information geometry for distinguishing the cause and effect with
relative entropy distances. CDCI~\cite{Duong_Nguyen_22Bivariate}
is based on the assumption that the conditional distribution of the
effect given the cause is invariant in shape.  Another interpretation
of this independence in the causal mechanism is the sum of the Kolmogorov
complexities~\cite{Kolmogorov_65Three} of $P_{X}$ and $P_{Y\mid X}$~\cite{Janzing_Scholkopf_10Causal,Marx_Vreeken_17Telling}.
The minimum description length (MDL) is adopted for approximating
the Kolmogorov complexity as this complexity is not computable. QCCD~\cite{Tagasovska_etal_20Distinguishing}
also follows this interpretation with non-parametric quantile regression
and code length computation for distinguishing the cause and effect.

\paragraph{}

%% file: preliminaries.tex
\subsection{Heteroscedastic Noise Models \& Causal-Effect Inference}

\textbf{\label{def:hnm}}(Heteroscedastic~Noise Model~\cite{Immer_etal_23Identifiablity}).\textbf{
}Given a random variable $X$ as the cause and a random variable $N_{Y}$
as the noise that is independent to $X$ ($N_{Y}\indep X$), a heteroscedastic
noise model or, in different terms, a location-scale noise model generates
the effect variable $Y$ with the following formulation:

\begin{equation}
Y:=f\left(X\right)+g\left(X\right)N_{Y},\label{eq:hnm}
\end{equation}
where $f:\mathcal{X}\rightarrow\mathbb{R}$ and $g:\mathcal{X}\rightarrow\mathbb{R}_{+}$.

Without loss of generality, for easier estimation, the mean of the
noise $\mathbb{E}\left[N_{Y}\right]$ is assumed to be equal to $0$.
The functions $f\left(X\right)$ and $g\left(X\right)$ contribute
to the mean and standard deviation of the noise. To estimate $f\left(X\right)$
and $g\left(X\right)$, we can estimate the mean and the standard
deviation of $p\left(Y\mid X\right)$. There are two approaches to
estimating the noise term $N_{Y}$ in HNM. GRCI~\cite{Strobl_Lasko_22Identifying}
uses leave-one-out cross-validation to learn the models for regressing
the mean and the mean absolute deviation (MAD). GRCI estimates MAD
instead of the standard deviation as the MAD can be estimated directly
and have smaller estimation errors. However, the use of cross-validation,
especially the leave-one-out method, has high a computational cost.
LOCI~\cite{Immer_etal_23Identifiablity}, on the other hand, models
the mean and the variance and uses the Gaussian log-likelihood $\log\mathcal{N}\left(Y\mid\mu\left(X\right),\sigma^{2}\left(X\right)\right)$
as the maximization objective. This approach allows the mean and the
variance can be optimized simultaneously, but the assumption of the
Gaussian likelihood is more restrictive compared to GRCI's approach. 

\subsection{Student's $t$-Distribution}

Student's $t$-distribution is a distribution that is considered as
the generalization of Gaussian distribution. Belonging to the location-scale
family as the Gaussian family, the $t$-distribution's parameters
also include a location $\mu$ and a scale $\tau^{2}$. In addition
to the location and scale, the third parameter of the $t$-distribution
is the degree of freedom $\nu$ ($\nu>0$). The probability density
function (PDF) of the $t$-distribution in defined as 
\begin{equation}
t\left(x\mid\mu,\tau^{2},\nu\right)=\frac{\Gamma\left(\frac{\nu+1}{2}\right)}{\Gamma\left(\frac{\nu}{2}\right)\sqrt{\pi\nu\tau^{2}}}\left(1+\frac{1}{\nu}\frac{\left(x-\mu\right)^{2}}{\tau^{2}}\right)^{-\frac{\nu+1}{2}},\label{eq:t-pdf}
\end{equation}
where $\Gamma\left(z\right)=\int_{0}^{\infty}t^{z-1}\exp\left(-t\right)dt$.
As derived from Eq.~(\ref{eq:t-pdf}), $t$-distribution is a heavy-tailed
distribution where the heaviness of the tails is controlled by the
degree of freedom $\nu$. As $\nu\rightarrow\infty$, the $t$-distribution
will become the Gaussian distribution. If $\nu>1$, the expected value
of the random variable $X\sim t\left(\mu,\tau^{2},\nu\right)$ is
defined as
\begin{equation}
\mathbb{E}\left[X\right]=\mu,\label{eq:t-mean}
\end{equation}
and when $\nu>2$, the variance of $X$ is definite and is calculated
from both the scale $\tau^{2}$ and the degree of freedom $\nu$ as
follows
\begin{equation}
\var\left[X\right]=\tau^{2}\frac{\nu}{\nu-2}.\label{eq:t-variance}
\end{equation}

Having an additional parameter $\nu$ controlling the variance allows
the $t$-distribution to handle sampling variability and extreme outliers
by adapting the shape of the distribution flexibly (see Figure~\ref{fig:teaser}
for an example). Therefore, the $t$-distribution is more robust than
the Gaussian distribution while maintaining the ability to approximate
the Gaussian distribution by using a high degree of freedom. Due to
these advantages, this distribution can be applied in place of the
Gaussian distribution improve the robustness such as in the Student-$t$
processes~\cite{Shah_etal_14Student,Tang_etal_17Student} as alternatives
for Gaussian processes, reliable variance networks estimation~\cite{Detlefson_etal_19Reliable},
or Student-$t$ variational autoencoder~\cite{Takahashi_etal_18Student}
for more robust density estimation. Following these intuitions, we
choose the $t$-distribution for the likelihood in our approach to
achieve a robust estimation of the heteroscedastic noise models.

%% file: method.tex
\subsection{Parameters Estimation with the $t$-Distribution}

Similar to previous approaches~\cite{Strobl_Lasko_22Identifying,Immer_etal_23Identifiablity},
we need to estimate the mean and the standard deviation to retrieve
the independent residual. We estimate these parameters via the likelihood
of the assumed effect $Y$ given the assumed cause $X$. The $t$-distribution
probability density function in Eq.~(\ref{eq:t-pdf}) is chosen for
this likelihood as follows
\begin{equation}
p\left(Y\mid X\right)=t\left(Y\mid\mu\left(X\right),\tau^{2}\left(X\right),\nu\left(X\right)\right),\label{eq:likelihood}
\end{equation}
and this likelihood is modeled using a neural network. Inspired by
the fact that the $t$-distribution is a Gaussian distribution with
the variance distributed as an inverse gamma distribution with two
parameters $a=\nu/2$ and $b=\nu\tau^{2}/2$. Instead of modeling
the scale function $\tau^{2}\left(X\right)$ and the degree of freedom
function $\nu\left(X\right)$ separately, we model two alternative
parameter functions $a\left(X\right)$ and $b\left(X\right)$. The
degree of freedom $\nu$ and the scale $\tau^{2}$ are then computed
as 
\begin{align}
\nu\left(X\right)=2a\left(X\right),\label{eq:model_df}\\
\tau^{2}\left(X\right)=\frac{b\left(X\right)}{a\left(X\right)}.\label{eq:model_scale}
\end{align}
By doing so, the scale $\tau^{2}$ depends on the degree of freedom
$\nu$, and will adapt accordingly to the change in the degree of
freedom. Since the degree of freedom must be greater than $2$ for
the variance to exist and the scale of the $t$-distribution must
be greater than $0$, the constraints $a\left(X\right)>1$ and $b\left(X\right)>0$
are applied when modeling. 

With this model, the neural network is optimized by minimizing the
negative log-likelihood as the loss function
\begin{align}
\mathcal{L}_{\text{NLL}} & =-\frac{1}{n}\sum_{i=1}^{n}\log t\left(y_{i}\mid\mu\left(x_{i}\right),\tau^{2}\left(x_{i}\right),\nu\left(x_{i}\right)\right)\label{eq:loss_nll}
\end{align}
where $n$ is the number of samples and $\left(x_{i},y_{i}\right)$
is the $i$-th sample from the dataset. There could be many models
that can be fitted with this loss function. The most preferable models
are the one with higher degrees of freedom which approach the Gaussian
distribution. To implement this preference, in addition to the negative
log-likelihood loss, we propose an additional constraint loss for
the degree of freedom $\nu\left(\cdot\right)$ as follows
\begin{align}
\mathcal{L}_{\text{constraint}} & =\frac{1}{n}\sum_{i=1}^{n}\left(\frac{\nu\left(x_{i}\right)}{\nu\left(x_{i}\right)-2}\right)^{2}\label{eq:loss_reg}
\end{align}
This constraint is chosen from the $t$-distribution's variance in
Eq.~(\ref{eq:t-variance}), where the variance is scaled from the
distribution's scale parameter $\tau^{2}$ with a factor of $\nu/\left(\nu-2\right)$.
The value of the constraint will decrease and approach $1$ when all
the values of $\nu\left(x_{i}\right)$ increase. Penalizing this constraint
will also reduce the scaling term of the variance in Eq.~(\ref{eq:t-variance}),
which makes the variance become more stable by having values closer
to $\tau^{2}$ instead of exploding to infinity. Moreover, as the
parameter $\tau^{2}$ in our approach is modeled with Eq.~(\ref{eq:model_scale}),
raising the value of $\nu$ which corresponds to $a$ will also decrease
the value of $\tau^{2}$. This behavior can avoid the underfitting
scenario where the location parameter $\mu$ is poorly predicted but
the scale is increased to account for the high error ranges. This
scenario also happens to the Gaussian log-likelihood where many approaches,
such as $\beta$-NLL~\cite{Seitzer_etal_22Pitfalls} or Faithful
Heteroscedastic~\cite{Stirn_etal_23Faithful}, has been proposed
to have more reliable mean regression by modifying the Gaussian log-likelihood
function, but these methods need to have a trade-off between the mean
accuracy and the log-likelihood. We do not have this trade-off since
the constraint term is a detached module and our method still tries
to find the models with the highest log-likelihood.

\subsection{Residual Estimation \& Cause-Effect Inference}

After estimating heteroscedastic noise models in Eq.~(\ref{eq:hnm}),
the residual can be computed from the estimands as follows
\begin{equation}
\tilde{N}_{Y}=\frac{Y-\tilde{f}\left(X\right)}{\tilde{g}\left(X\right)},\label{eq:residual}
\end{equation}
where $\tilde{N}_{Y}$, $\tilde{f}\left(X\right)$ and $\tilde{g}\left(X\right)$
denote the estimated $N_{Y}$, $f\left(X\right)$, and $g\left(X\right)$.
In our case with the $t$-distribution, the function $f\left(X\right)$
corresponds to $\mathbb{E}\left[Y\mid X\right]=\mu\left(X\right)$
and $g^{2}\left(X\right)$ corresponds to $\var\left[Y\mid X\right]=\tau^{2}\left(X\right)\frac{\nu\left(X\right)}{\nu\left(X\right)-2}$.
After learning the model, the estimated residual can be determined
from the estimated parameters of the $t$-distribution as follows
\begin{align}
\tilde{N}_{Y} & =\frac{Y-\tilde{\mu}\left(X\right)}{\sqrt{\tilde{\tau^{2}}\left(X\right)\frac{\tilde{\nu}\left(X\right)}{\tilde{\nu}\left(X\right)-2}}}\label{eq:residual_t}
\end{align}
where $\tilde{\mu}$, $\tilde{\tau^{2}}$, $\tilde{\nu}$, $\tilde{a}$,
and $\tilde{b}$ are the parameters estimated with the $t$-distribution
likelihood. 

As the noise term $N_{Y}$ and the cause $X$ are assumed to be independent,
the dependence between the estimated noise $\tilde{N}_{Y}$ and the
cause $X$ quantified using any non-parametric independent test can
be utilized to support the cause-effect inference step. In previous
approaches, GRCI~\cite{Strobl_Lasko_22Identifying} estimates the
the independence between $\tilde{N}_{Y}$ and $X$ via the mutual
information, whereas LOCI~\cite{Immer_etal_23Identifiablity} uses
the Hilbert-Schmidt independence criterion (HSIC)~\cite{Gretton_etal_05Measuring}.
The Gaussian log-likelihood is also proposed in LOCI~\cite{Immer_etal_23Identifiablity}
as a score for quantifying the causal direction. However, using Gaussian
likelihood scoring can cause model misspecification when the distribution
of the noise is non-Gaussian, especially in heteroscedastic noise
models~\cite{Schultheiss_Buhlmann_23Pitfalls}. A more suitable choice
should be using the quantified dependence between $\tilde{N}_{Y}$
and $X$. In this work, we choose the HSIC score~\cite{Gretton_etal_05Measuring}
for easier comparison with LOCI~\cite{Immer_etal_23Identifiablity},
which chooses the Gaussian likelihood for estimating the heteroscedastic
noise models. Higher HSIC score indicates that two variables have
a higher degree of independence.

%% file: experiment.tex
\begin{figure*}[!t]
\centering{}\emph{\includegraphics[width=0.8333\textwidth]{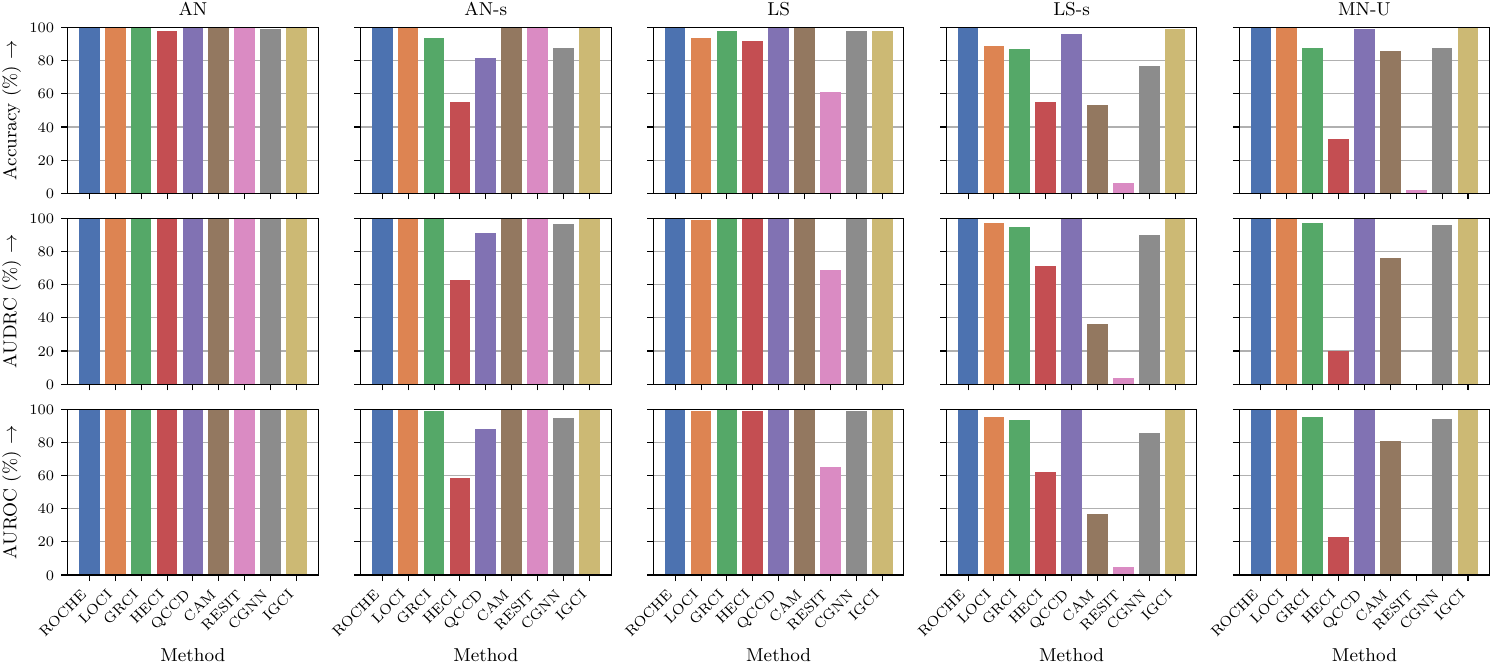}}\caption{\label{fig:results_simple}Performance on synthetic additive (AN \&
AN-s), location-scale (LS \& LS-s), and multiplicative (MN-U) noise
models. Higher accuracy, AUDRC, and AUROC are preferable. Our ROCHE
approach is compared with methods for heteroscedastic noise models:
LOCI~\cite{Immer_etal_23Identifiablity}, GRCI~\cite{Strobl_Lasko_22Identifying},
HECI~\cite{Xu_etal_22Inferring}, and QCCD~\cite{Tagasovska_etal_20Distinguishing},
and methods not designed for heteroscedasticity: CAM~\cite{Buhlmann_etal_14CAM},
RESIT~\cite{Peters_etal_14Causal}, CGNN~\cite{Goudet_etal_18Learning},
and ICGI~\cite{Janzing_etal_12Information}. Our ROCHE approach achieves
perfect results on all 5 benchmark datasets.}
\end{figure*}

\begin{figure*}[t]
\begin{centering}
\includegraphics[width=0.6667\textwidth]{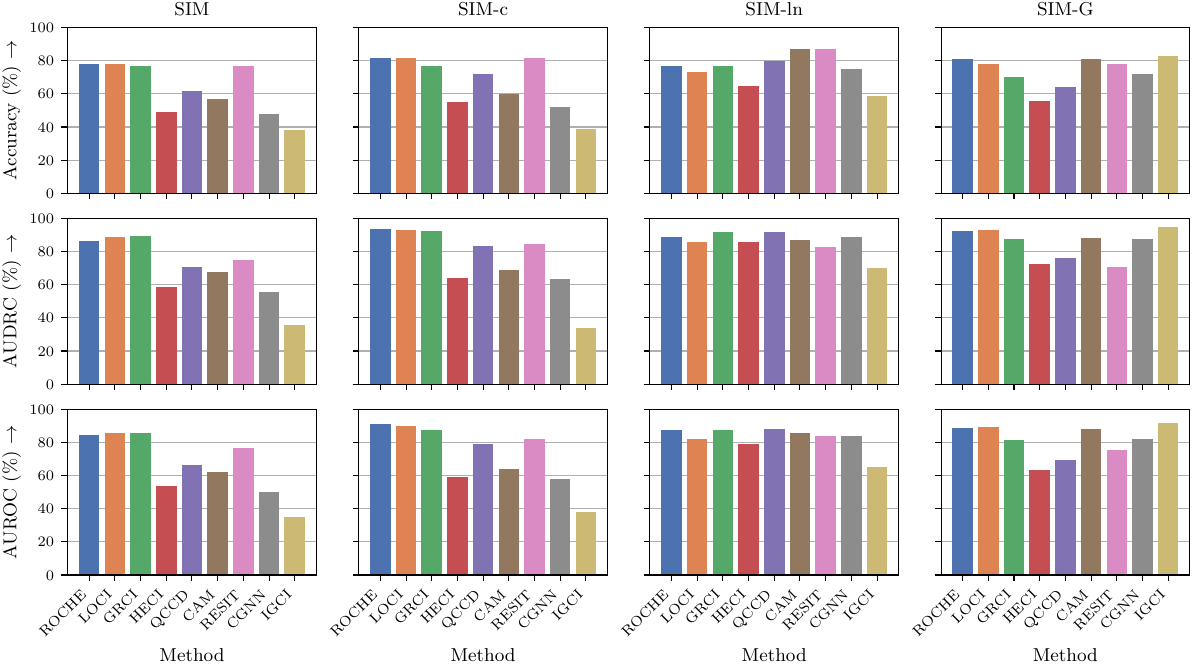}\caption{\label{fig:results_complex1}Performance on SIM, SIM-c, SIM-ln, and
SIM-G datasets. Higher accuracy, AUDRC, and AUROC are preferable.
Our ROCHE approach is compared with methods for heteroscedastic noise
models: LOCI~\cite{Immer_etal_23Identifiablity}, GRCI~\cite{Strobl_Lasko_22Identifying},
HECI~\cite{Xu_etal_22Inferring}, and QCCD~\cite{Tagasovska_etal_20Distinguishing},
and methods not designed for heteroscedasticity: CAM~\cite{Buhlmann_etal_14CAM},
RESIT~\cite{Peters_etal_14Causal}, CGNN~\cite{Goudet_etal_18Learning},
and ICGI~\cite{Janzing_etal_12Information}. ROCHE acquires comparable
and stable results on most datasets and accomplishes better results
in comparison with the Gaussian-likelihood-based LOCI. Regression-based
methods with subsequent independence tests including ROCHE, LOCI,
GRCI, and RESIT have better performance in the occurrence of confounders.}
\par\end{centering}
\end{figure*}

\subsection{Experimental Settings}

In this section, we compare our ROCHE approach with related bivariate
causal discovery methods designed for heteroscedastic noise models
and homoscedastic noise models to demonstrate the effectiveness of
our approach. 

\paragraph{Baselines}

Evaluated state-of-the-art causal discovery methods for heteroscedastic
noise models including QCCD~\cite{Tagasovska_etal_20Distinguishing},
HECI~\cite{Xu_etal_22Inferring}, GRCI~\cite{Strobl_Lasko_22Identifying},
and LOCI~\cite{Immer_etal_23Identifiablity}. Among these approaches,
the most related method to our approach is LOCI which employs the
Gaussian likelihood for estimating the location and scale of the model.
For LOCI, we use the neural network version with HSIC as the score
for easier comparisons. For methods that are designed for homoscedastic
or additive noise models, we use CAM~\cite{Buhlmann_etal_14CAM}
and RESIT~\cite{Peters_etal_14Causal}. For other baselines that
do not assume heteroscedastic noise models, we consider ICGI~\cite{Janzing_etal_12Information}
with Gaussian reference measures and CGNN~\cite{Goudet_etal_18Learning}.

\paragraph{Datasets}

We use both synthetic and real causal discovery benchmarks to evaluate
our approach. 

The first group of chosen benchmarks is the five synthetic datasets
from~\cite{Tagasovska_etal_20Distinguishing} including AN, AN-s,
LS, LS-s, and MN-U. These five datasets are generated from structural
causal models that are heteroscedastic (or location-scale) noise models
(LS and LS-s datasets) and special cases of heteroscedastic noise
models consisting of homoscedastic additive noise models (AN and AN-s
datasets) and multiplicative noise models (MN-U). The datasets denoted
with ``-s'' and the MN-U dataset use invertible sigmoid-type functions
for the generative functions which can make the identification of
the causal direction more complicated. 

In the second group of benchmarks, we consider more difficult synthetic
datasets comprising the SIM, SIM-c, SIM-ln, and SIM-G from~\cite{Mooij_etal_16Distinguishing}
and the Multi and Net datasets from~\cite{Goudet_etal_18Learning},
the Cha dataset from~\cite{Guyon_etal_19Cause}. The four SIM, SIM-c,
SIM-ln, SIM-G datasets are simulated with the functions $X:=f_{X}\left(N_{X}\right)$
and $Y:=f_{Y}\left(X,N_{Y}\right)$ in the cases without a confounder
(in SIM, SIM-ln, and SIM-G) and the functions $Z:=f_{Z}\left(N_{Z}\right)$,
$X:=f_{X}\left(N_{X},N_{Z}\right)$, and $Y:=f_{Y}\left(X,N_{Y},N_{Z}\right)$
in the cases with a confounder $Z$ (in SIM-c). In the SIM-ln dataset,
low levels of noise are applied in the models, and the SIM-G dataset
has approximations of Gaussian distributions for the cause $X$ and
approximately Gaussian non-linear additive noise generative models.
The Multi dataset is generated with pre-additive noises ($Y:=f\left(X+N_{Y}\right)$),
post-additive noises (similar to conventional additive noise models),
pre-multiplicative noise ($Y:=f\left(X\times N_{Y}\right)$), and
post-multiplicative noise ($Y:=f\left(X\right)\times N_{Y}$). The
pairs in the Net dataset are generated with neural networks with random
distribution for $X$, such as exponential, gamma, log-normal, or
Laplace distribution. The 300 continuous variable pairs in the Cha
benchmark are chosen from the ChaLearn Cause-Effect Pairs Challenge~\cite{Guyon_etal_19Cause}. 

For a real-world benchmark, we choose the Tübingen cause-effect pairs~\cite{Mooij_etal_16Distinguishing}
for comparison. This dataset contains pairs of cause(s) and effect(s)
gathered from many sources with diverse domains. Each pair in this
benchmark has a corresponding weight to scale down the results of
pairs having similar properties. For that reason, the evaluation results
for this benchmark are also weighted according to these weights.

\paragraph{Metrics}

The cause-effect inference results are assessed with three common
metrics in bivariate causal discovery literature consisting of the
accuracy score, the area under the decision rate curve (AUDRC), and
the area under the receiver operating characteristic curve (AUROC).
The accuracy score is the simplest evaluation metric which computes
the percentage of the pairs in the datasets whose causal directions
are correctly predicted. The AUDRC criterion is proposed in~\cite{Immer_etal_23Identifiablity}
for quantifying the extent to which decision confidence aligns with
accuracy. Having higher AUDRC means that the methods will predict
the correct causal directions if the confidence of the scores is high,
and the incorrect cases are caused by lower confidence of the predicted
score. The last chosen metric is the AUROC score. This score indicates
the ability of the methods to predict correctly without intermingling
between the two causal directions. To compensate for the class imbalance
when computing the AUROC score, we augment the results by adding an
additional result for the reversed direction with a reversed label
for each result acquired.

\begin{figure}[!t]
\begin{centering}
\emph{\includegraphics[width=1\columnwidth]{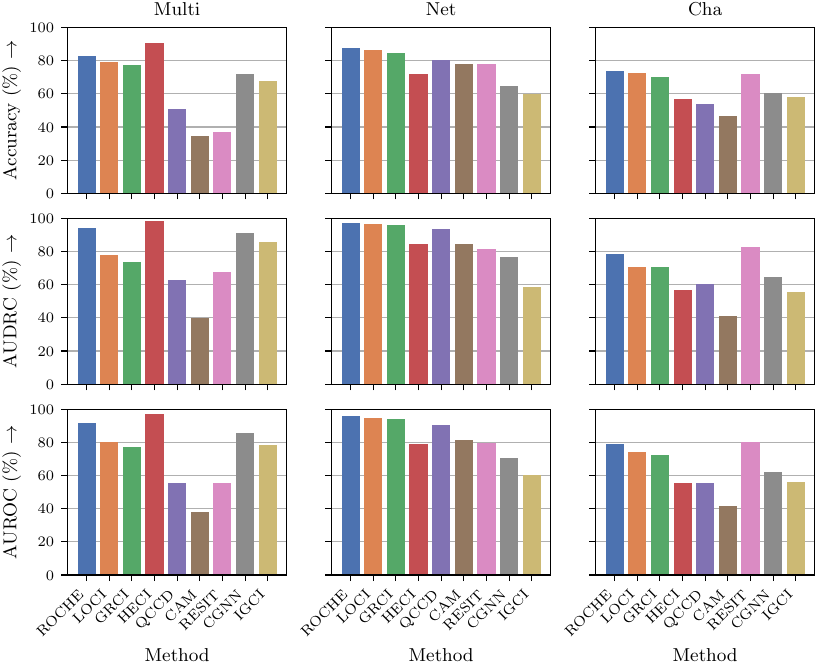}}
\par\end{centering}
\centering{}\caption{\label{fig:results_complex2}Performance on Multi, Net, and Cha datasets.
Higher accuracy, AUDRC, and AUROC are preferable. Our ROCHE approach
is compared with methods for heteroscedastic noise models: LOCI~\cite{Immer_etal_23Identifiablity},
GRCI~\cite{Strobl_Lasko_22Identifying}, HECI~\cite{Xu_etal_22Inferring},
and QCCD~\cite{Tagasovska_etal_20Distinguishing}, and methods not
designed for heteroscedasticity: CAM~\cite{Buhlmann_etal_14CAM},
RESIT~\cite{Peters_etal_14Causal}, CGNN~\cite{Goudet_etal_18Learning},
and ICGI~\cite{Janzing_etal_12Information}. ROCHE acquires the best
results on the Net dataset and the second highest scores on Multi
and datasets.}
\end{figure}
\begin{figure}[t]
\begin{centering}
\emph{\includegraphics[width=1\columnwidth]{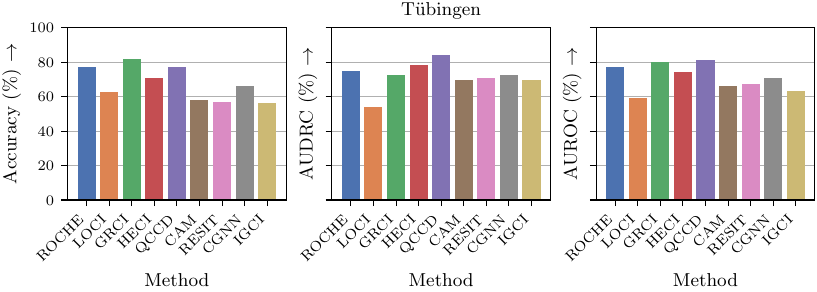}}\caption{\label{fig:results_tueb}Performance on Tübingen cause-effect pairs
datasets. Higher accuracy, AUDRC, and AUROC are preferable. Our ROCHE
approach is compared with methods for heteroscedastic noise models:
LOCI~\cite{Immer_etal_23Identifiablity}, GRCI~\cite{Strobl_Lasko_22Identifying},
HECI~\cite{Xu_etal_22Inferring}, and QCCD~\cite{Tagasovska_etal_20Distinguishing},
and methods not designed for heteroscedasticity: CAM~\cite{Buhlmann_etal_14CAM},
RESIT~\cite{Peters_etal_14Causal}, CGNN~\cite{Goudet_etal_18Learning},
and ICGI~\cite{Janzing_etal_12Information}. ROCHE provides promising
results and performs significantly better than LOCI with Gaussian
likelihood estimation. In general, methods with heteroscedasticity
have better performance compared to methods that do not assume heteroscedastic
noise models.}
\par\end{centering}
\end{figure}

\subsection{Performance on Synthetic Additive, Location-Scale, and Multiplicative
Noise Datasets}

Figure~\ref{fig:results_simple} illustrates the results of all considered
approaches on the AN, AN-s, LS, LS-s, and MN-U datasets. Our ROCHE
approach achieves the perfect results in all of these datasets which
demonstrate our model's ability to estimate effectively synthesized
heteroscedastic noise data, and special cases including homoscedastic/additive
noise and multiplicative noise data. LOCI, which uses neural networks
with Gaussian likelihood estimation, has some misspecified cases in
LS and LS-s datasets. One case where the causal direction is wrongly
predicted has already been presented in Figure~\ref{fig:teaser}.
The reason for these misspecified pairs is due to the suboptimal fit
of the variance in regions with low sample densities. Other methods
designed for heteroscedasticity, which are GRCI, HECI, and QCCD, have
some pairs misidentified, especially when the generative functions
are invertible on AN-s, LS-s, and MN-U datasets. CAM and RESIT achieve
perfect scores on the assumption-satisfied additive noise datasets
but as the datasets become more difficult these methods cannot perform
stably. With no assumption in the structural model, CGNN performs
well on AN and LS datasets. However, this approach is also affected
by invertible generative functions. Despite not being designed for
heteroscedastic noise models, IGCI has nearly perfect results in all
cases.

\subsection{Performance on More Complex Synthetic Datasets}

The results for the 7 remaining simulated datasets are presented in
Figures~\ref{fig:results_complex1} and~\ref{fig:results_complex2}.
These results show that our approach can adapt better to diverse model
configurations and noise distributions compared to other approaches.
On SIM, SIM-ln, and SIM-G datasets, our ROCHE's results are the best
ones in the group of methods with heteroscedastic design. On confounding
cases in the SIM-c dataset, our method is more robust and retrieve
the causal directions more accurately than other current models. In
most cases, the Student's $t$-distribution estimation approach in
ROCHE provides better results compared to the Gaussian estimation
in LOCI and the conditional mean and condition mean absolute deviation
estimation in GRCI. Noticeably, regression-based methods that regress
the residuals and use independence testing for determining causal
direction, such as ROCHE, LOCI, GRCI, and RESIT, have demonstrated
superior performance in the scenario when confounding variables are
present in the SIM-c dataset. Despite having an assumption of low
noise levels, IGCI has the lowest distinguishing performance on the
SIM-ln dataset. Our approach obtains the second highest results on
the Multi and Cha datasets and the highest results on the Net dataset.
QCCD predicts decently on the Net and SIM-ln datasets. But on other
datasets, the results are less stable with nearly random predictions
on the Cha dataset. Although HECI has the best scores on the Multi
datasets, this method does not perform as well on the remaining datasets.
Homoscedastic methods---CAM and RESIT have high variances in the
results among different benchmarks with even sub-standard results
on the Multi and Cha datasets. CGNN also exhibits high discrepancy
between datasets with decent results in most datasets, except for
the SIM and SIM-c benchmarks. On this group of complex benchmarks,
our approach has the most stable performance across these diverse
benchmark datasets. 

\begin{table}[!t]
\caption{\label{tab:results_overall}Overall performance on 13 benchmarks.
Reported results are the mean$\pm$std of the~Accuracy,~AUDRC,~and~AUROC
across the benchmarks. Higher Accuracy, AUDRC, and AUROC are preferable.
The best results are presented in \textbf{bold} style. Our ROCHE approach
achieves the best overall results with high stability, which demonstrates
the robustness of our method.}

\centering{}\input{figs/overall.tex}
\end{table}

\subsection{Performance on Real-World Data}

In addition to synthetic datasets, we also evaluate our method on
the real-world Tübingen cause-effect pairs dataset. The accuracy,
AUDRC, and AUROC results on this benchmark are depicted in Figure~\ref{fig:results_tueb}.
Utilizing the more robust $t$-distribution for the likelihood in
ROCHE instead of the Gaussian distribution in LOCI contributes to
a huge performance gain on the Tübingen benchmark. Our approach accomplishes
promising results with the second highest in accuracy and the third
highest in other evaluation metrics on this dataset which are satisfactory
and comparable to other methods. The lower robustness of the Gaussian
likelihood estimation on real data affects the performance of LOCI
substantially and leads to the method's lowest AUDRC and AUROC in
this benchmark. On the more complicated Tübingen dataset, methods
targeted heteroscedastic noise models provide better predictions compared
to non-heteroscedastic baselines, especially in the case of CAM and
RESIT. 

\subsection{Overall Performance across All Benchmarks}

We summarize the overall performance across 13 benchmarks in Table~\ref{tab:results_overall}
to have a complete evaluation for all methods. In general, our method
has the best average results in all three evaluation metrics with
the percentage values of accuracy, AUDRC, and AUROC being $87.69\%$,
$92.86\%$, and $92.02\%$ respectively. In addition, ROCHE is more
robust in comparison with remaining methods with margins of errors
less than $10\%$ in AUDRC and AUROC criteria. The standard deviations
of our approach have the second lowest value of $10.23\%$ for the
accuracy score, which is slightly higher than the best value of $9.44\%$
from GRCI, and the lowest values of $8.16\%$ and $7.90\%$ for AUDRC
and AUROC scores respectively.

%% file: figs/overall.tex
\resizebox{\columnwidth}{!}{
\begin{tabular}{l|c|c|c}
\toprule 
Method & Accuracy (\%) $\uparrow$ & AUDRC (\%) $\uparrow$ & AUROC (\%) $\uparrow$\tabularnewline
\midrule
\textbf{ROCHE (Ours)} & $\mathbf{87.69\pm10.23}$ & $\mathbf{92.86\pm\phantom{0}8.16}$ & \textbf{$\mathbf{92.02\pm\phantom{0}7.90}$}\tabularnewline
LOCI~\cite{Immer_etal_23Identifiablity} & $84.24\pm11.43$ & $89.04\pm13.42$ & $88.58\pm11.66$\tabularnewline
GRCI~\cite{Strobl_Lasko_22Identifying} & $83.20\pm\phantom{0}9.44$ & $89.73\pm10.19$ & $88.86\pm\phantom{0}8.70$\tabularnewline
HECI~\cite{Xu_etal_22Inferring} & $65.25\pm18.17$ & $73.34\pm21.19$ & $69.60\pm20.97$\tabularnewline
QCCD~\cite{Tagasovska_etal_20Distinguishing} & $78.21\pm16.61$ & $85.77\pm13.70$ & $82.69\pm15.79$\tabularnewline
CAM~\cite{Buhlmann_etal_14CAM} & $72.42\pm21.23$ & $73.89\pm21.97$ & $72.77\pm22.33$\tabularnewline
RESIT~\cite{Peters_etal_14Causal} & $64.45\pm30.48$ & $68.36\pm30.05$ & $67.08\pm29.96$\tabularnewline
CGNN~\cite{Goudet_etal_18Learning} & $73.89\pm15.43$ & $83.35\pm14.48$ & $79.87\pm15.55$\tabularnewline
IGCI~\cite{Janzing_etal_12Information} & $73.69\pm22.98$ & $77.26\pm24.04$ & $76.00\pm23.54$\tabularnewline
\bottomrule
\end{tabular}}

%% file: conclusion.tex
In this work, we have presented ROCHE, a robust estimation approach
for bivariate causal discovery with heteroscedastic noise models.
Instead of relying on the Gaussian distribution, as done in previous
approaches which may not accurately capture the model parameters,
we adopt the more robust Student's $t$-distribution. This choice
allows us to account for sampling variability and extreme values more
effectively. We propose a framework for modeling the parameters using
the $t$-distribution along with necessary constraints to ensure the
resulting estimators are suitable for subsequent cause-effect inference
tasks. The effectiveness of ROCHE has been demonstrated through comprehensive
experiments, yielding promising results and showcasing the best overall
performance across 13 benchmark datasets. Furthermore, our approach
exhibits robustness, as evidenced by the lowest overall margins of
errors across various evaluation metrics. In future work, we aim to
extend our approach to handle multivariate heteroscedastic noise models,
thus enhancing its applicability to real-world data scenarios.